\def\year{2023}\relax
\documentclass[letterpaper]{article} 
\usepackage{aaai22}  
\usepackage{times}  
\usepackage{helvet}  
\usepackage{courier}  
\usepackage[hyphens]{url}  
\usepackage{graphicx} 
\usepackage{natbib}  
\usepackage{caption} 
\DeclareCaptionStyle{ruled}{labelfont=normalfont,labelsep=colon,strut=off} 
\usepackage{algorithm}
\usepackage{algorithmic}
\usepackage{multirow}
\usepackage[table,xcdraw]{xcolor}
\usepackage{hyperref}
\usepackage{newfloat}
\usepackage{listings}
\floatstyle{ruled}
\newfloat{listing}{tb}{lst}{}
\floatname{listing}{Listing}

\title{Deep Convolutional Neural Networks for Palm Fruit Maturity Classification\thanks{ The code is available at \href{https://github.com/mingqianghan/PalmFruitMaturity-CNN}{GitHub}.}}

\author {
    Mingqiang Han,\textsuperscript{\rm 1}
    Chunlin Yi, \textsuperscript{\rm 2}
}
\affiliations {
    \textsuperscript{\rm 1} Department of Biological and Agricultural Engineering, Kansas State University\\
    \textsuperscript{\rm 2} Department of Electrical and Computer Engineering, Kansas State University\\
    \{mingqiang, cyi\}@ksu.edu
}

\usepackage{bibentry}
\begin{document}

\maketitle

\begin{abstract}
To maximize palm oil yield and quality, it is essential to harvest palm fruit at the optimal maturity stage. This project aims to develop an automated computer vision system capable of accurately classifying palm fruit images into five ripeness levels. We employ deep Convolutional Neural Networks (CNNs) to classify palm fruit images based on their maturity stage. A shallow CNN serves as the baseline model, while transfer learning and fine-tuning are applied to pre-trained ResNet50 and InceptionV3 architectures. The study utilizes a publicly available dataset of over 8,000 images with significant variations, which is split into 80\% for training and 20\% for testing. The proposed deep CNN models achieve test accuracies exceeding 85\% in classifying palm fruit maturity stages. This research highlights the potential of deep learning for automating palm fruit ripeness assessment, which can contribute to optimizing harvesting decisions and improving palm oil production efficiency.
\end{abstract}

\section{Introduction}

Palm fruit is a high economic value plantation crop and one of the most widely used vegetable oils. Its cultivation is an important strategic agricultural industry in some tropical countries, such as Egypt, Iran, and Indonesia. Crude palm oil accounts for 34.2\% of the world's largest vegetable oil consumption due to its modifiable chemical composition and suitability in various food applications \cite{hcrt:83}. There is a constant demand for the production of high-quality palm oil with lower production costs \cite{hcrt:84}. The quality of palm oil produced and the overall marketability depend heavily on the ripeness level of the palm fruit. One of the biggest challenges in the processing of fruits during oil production is the classification of fresh oil palm fruits for their maturity. Hence, to maximize the yield and quality of palm oil, it is crucial to harvest oil palms at their correct ripeness stage.

The conventional method for grading palm fruit maturity relies on manual detection, requiring experts to visually sort the fruit based on texture, shape, and color. However, this manual process is labor-intensive and inefficient, often leading to bias and human error. To enhance productivity, an automated harvesting system should be developed. Recently, some fruit harvesting robots have been introduced to improve yield and quality while reducing production costs and delays \cite{c:83}. The key capability of these robots is their ability to interpret and analyze collected data. To achieve optimal performance, they must be equipped with an accurate sorting system that classifies and assesses fruit maturity in real time.

The practical success of designing such maturity sorting systems still requires further research due to the challenges posed by unstructured and unconstrained agricultural environments \cite{hcrt:85}. The most practical approach for an automatic system is to use an Artificial Intelligence system that can classify the ripeness of palm fruit from an image. In this study, we aim to develop a computer vision system to detect ripeness stages. This system, installed on a harvesting robot, can accurately analyze and classify fruits based on different types and ripeness levels in real time, providing references for decision-making. 

The remainder of this paper is organized as follows. Section 2 reviews related work. Section 3 provides a detailed description of the proposed method. Section 4 presents the experimental results. Finally, Section 5 concludes the paper.

\begin{figure*}[t]
\centering
\includegraphics[width=0.96\textwidth]{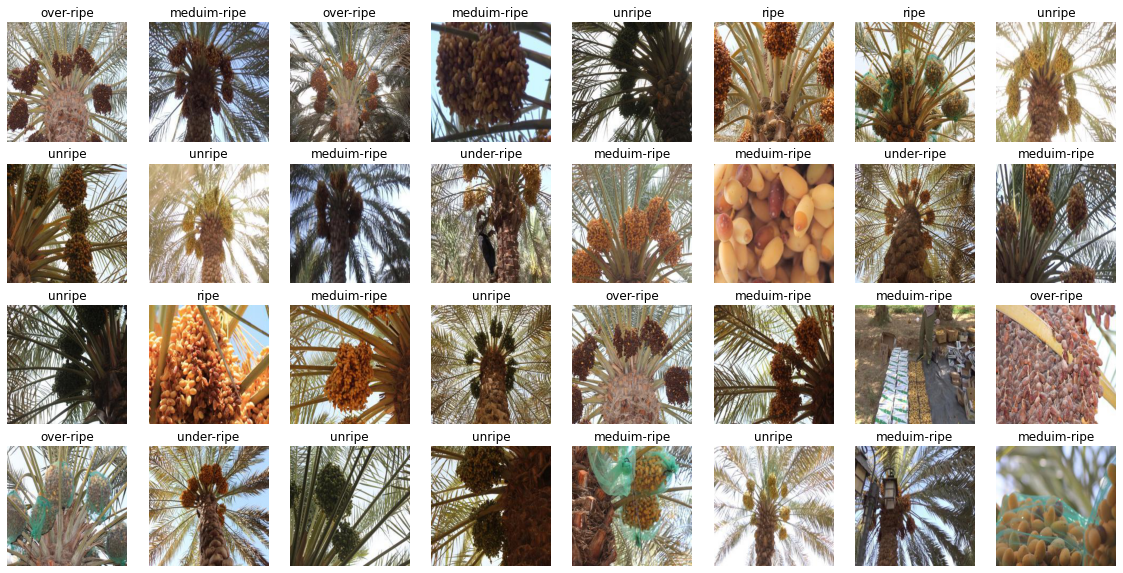} 
\caption{Sample images from the dataset are grouped into five ripeness classes: unripe, under-ripe, medium-ripe, ripe, and overripe. These images exhibit significant variations in scale, angle, and illumination, with some palm bunches partially covered by green bags.}
\label{fig1}
\end{figure*}

\section{Related Work}

\subsection{Methods for Classifying Palm Fruit Ripeness}
To ensure that the maximum amount of palm oil is extracted from the fruit, it is important to harvest it at the correct stage of ripeness. As the fruit enters the under-ripe stage, its oil content begins to increase. Ripe fruit has a reddish-orange color and, at the same time, the highest oil content. However, once the fruit becomes overripe, its fatty acid content starts to increase, reducing the oil quality. In recent years, various automated fruit grading systems have been proposed and tested. These methods can be classified into two categories. One approach relies on image processing techniques, which extract color features by preprocessing palm fruit images. The other employs CNNs, which automatically learn features from data.

In the traditional method, features from RGB and HSI color models are extracted from images and used as predictors for classification. Color histogram, color moment, and color correlogram \cite{hcrt:86} are derived from the original images as predictive features. RGB values and color indices, obtained from RGB color images through stepwise discriminant analysis, have been used for maturity determination \cite{hcrt:87}. Additionally, the mean and range of RGB digital values have been applied to determine fruit maturity \cite{hcrt:88}. To enhance feature extraction, dimensionality reduction techniques such as principal component analysis (PCA) \cite{cc:83} have been employed. The extracted features are then fed into classification algorithms, including k-nearest neighbors (k-NN) \cite{cc:84}, Support Vector Machine (SVM) \cite{hcrt:89}, and Multi-Layer Perceptron (MLP) \cite{hcrt:88}. However, the results obtained with this method are not reliable because ambient lighting significantly affects the values of these color models, thereby impacting the extracted features. Moreover, the camera configuration must remain fixed, which is impractical in real-world applications.

In recent years, deep learning algorithms have demonstrated remarkable performance in image processing tasks. The limitations of traditional methods can be mitigated by using deep learning techniques. AlexNet \cite{hcrt:90} was the first deep learning architecture applied to palm ripeness classification. This study utilized the pre-trained CNN model, AlexNet, on 120 images of palm oil fruit, with 30 images from each of the four ripeness levels. A more recent study \cite{cc:85} compared the performance of AlexNet and DenseNet. Both models were trained and tested using a dataset of 400 palm oil fruit images, with 60\% allocated for training, 20\% for validation, and 20\% for testing. The test accuracy was 77\% for AlexNet and 85\% for DenseNet. However, the number of images used in both studies was very limited, and all palm fruit images were captured post-harvest. As a result, the findings are not reliable, and the models cannot be effectively applied to the practical harvesting process.

To develop a more robust vision system, a larger dataset was used for palm fruit bunch maturity classification \cite{hcrt:91}. This dataset consisted of more than 8,000 images of fruit bunches from five palm types at various pre-maturity and maturity stages. The study applied AlexNet and VGGNet architectures, achieving test accuracies of 93.36\% and 95.78\%, respectively. However, in this study, the number of test samples was larger than the training samples. Upon reviewing their work, we found that a significant number of similar images were present in both the training and test sets. Consequently, the high test accuracy depended heavily on the specific data split. If the dataset were randomly shuffled before splitting into training and test sets, similar results would not be achievable. Thus, further studies are needed to explore the feasibility of applying deep learning algorithms to practical palm bunch ripeness classification tasks.

\begin{figure*}[t]
\centering
\includegraphics[width=0.9\textwidth]{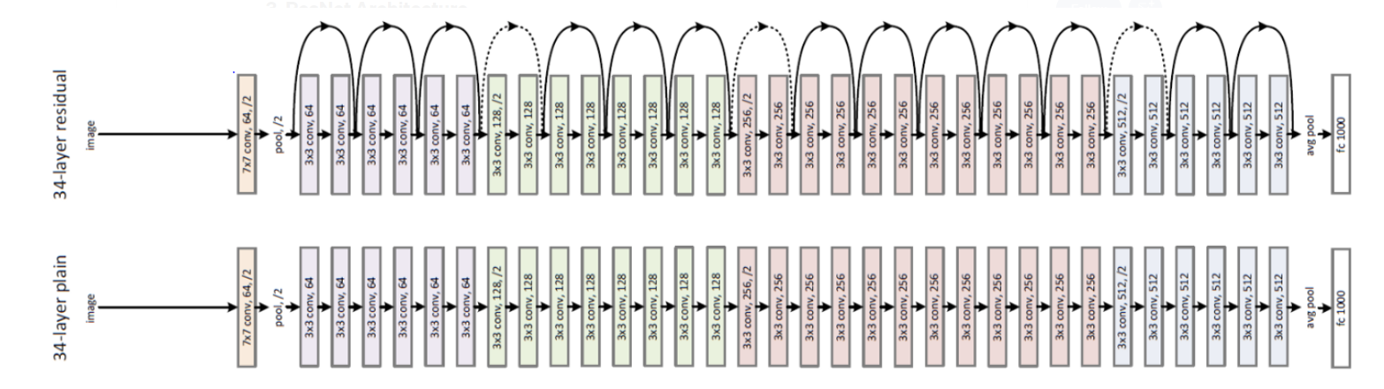} 
\caption{ The ResNet architecture }
\label{ResNet}
\end{figure*}

\begin{figure*}[t]
\centering
\includegraphics[width=0.9\textwidth]{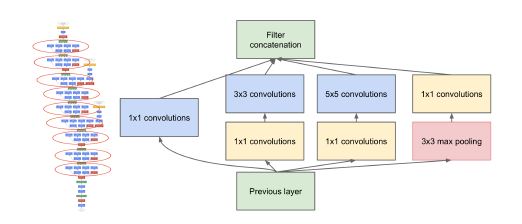} 
\caption{ The InceptionV3 architecture and the Inception module }
\label{incpt}
\end{figure*}

\subsection{Dataset}
Although weaknesses exist in the previous work \cite{hcrt:91}, the dataset used \cite{hcrt:92} is valuable and meaningful for palm research. To the best of our knowledge, this is the only publicly available dataset for palm fruit pre-harvesting and harvesting applications. This dataset can support various applications in both the pre-harvesting and harvesting stages. The first dataset, designed for palm type and maturity level classification, consists of 8,079 color images, each with a resolution of $224 \times 224$, captured from 29 palms. These images represent seven palm fruit maturity stages and exhibit a high degree of variation, reflecting the challenges of real-world environments in orchards. Variations in the dataset include different viewing angles and scales, varying daylight conditions, and the presence of palm bunches covered by protective bags. For multi-scale analysis, the images include partial bunches, whole bunches, and multiple bunches within a single frame. To simulate real-world harvesting conditions, images were captured under different natural daylight settings, specifically in the morning (9:00–11:00) and afternoon (3:00–5:00). Additionally, some images were taken under poor illumination conditions from different camera angles relative to the sun. The dataset covers five palm varieties, including some high-quality varieties where bunches are covered with protective bags to shield them from dust, pests, and rain. In particular, green bags were used to cover bunches after reaching the medium-ripe stage. Overall, this dataset is comprehensive and diverse enough to support the development of a robust vision system for palm fruit maturity classification and harvesting applications.

\begin{figure}[t]
\centering
\includegraphics[width=0.4\columnwidth]{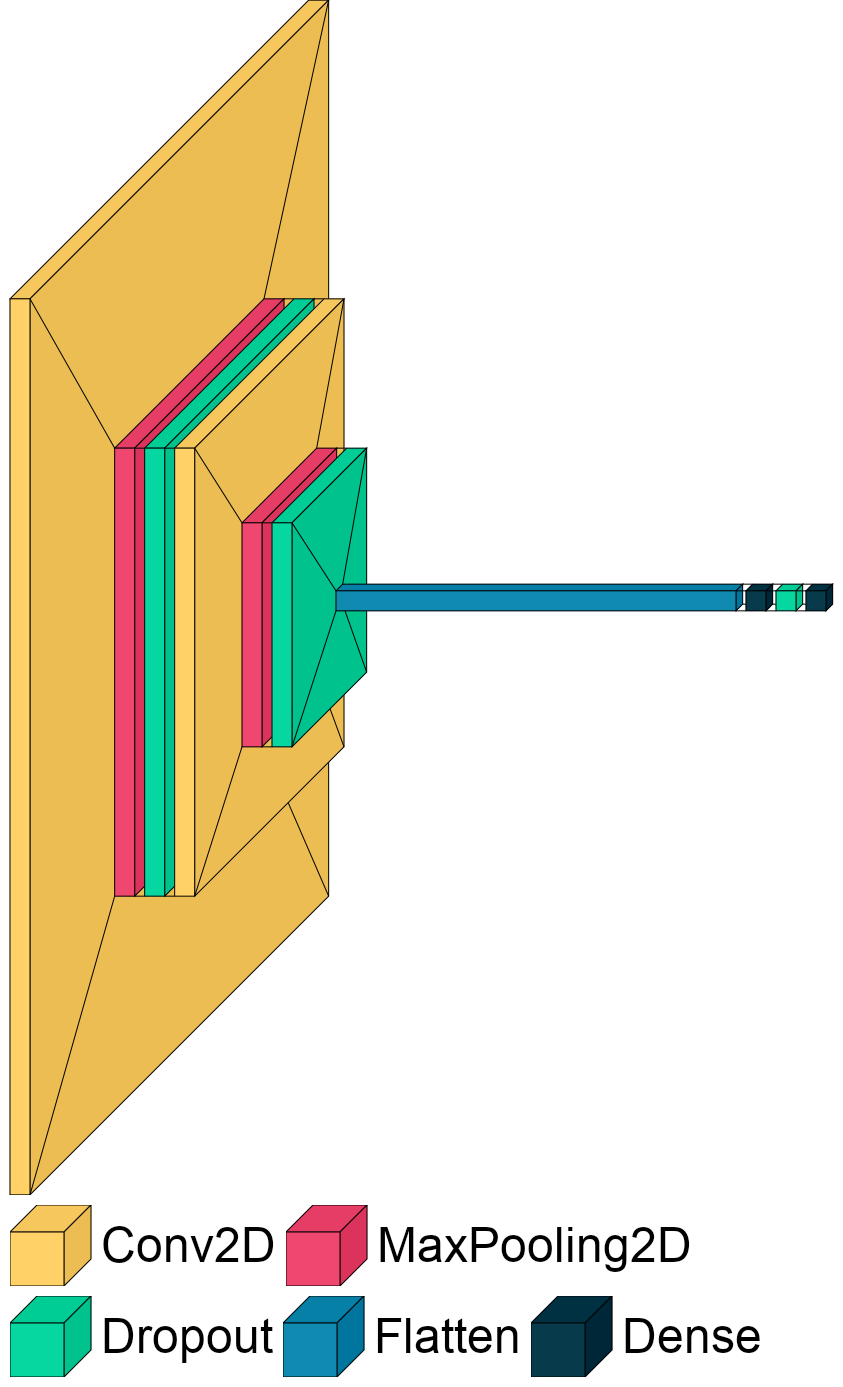} 
\caption{ Summary of the baseline model architecture.}
\label{fig_base}
\end{figure}

\subsection{Our Methodology}
In this study, we utilize deep CNNs to classify palm fruit ripeness levels, leveraging their strong capability for automatic feature extraction in this classification task. We apply transfer learning with fine-tuning using two pre-trained CNN models: ResNet50 and InceptionV3. To standardize classification, we combine certain ripeness levels from the published dataset \cite{hcrt:92} into five categories: unripe, under-ripe, medium-ripe, ripe, and overripe. Sample images representing these five ripeness levels are shown in Figure \ref{fig1}. The dataset was split into 80\% for training and 20\% for testing.

\section{Proposed Method}
This study explores three classification models for determining palm fruit maturity. The input consists of a stream of images captured by an RGB video camera in an orchard, as described in \cite{hcrt:92}. The output is the predicted maturity stage of the palm fruit in each image. For maturity classification, which is a multi-class classification task, we employ a shallow CNN as a baseline model. Additionally, we explore transfer learning with fine-tuning using two pre-trained CNN models: ResNet50 and InceptionV3, both of which are widely used for image classification tasks.

\subsection{CNN Deep Learning }
Image classification is one of the core challenges in computer vision, with a wide range of practical applications. Significant attention has been given to neural networks, particularly CNNs. CNNs utilize deep convolutional layers and non-linearity to automatically learn local and spatial features directly from images, eliminating the need for manual feature extraction. However, CNNs typically require a large amount of labeled training data for learning weight parameters. Additionally, their high computational cost necessitates the use of powerful GPUs to accelerate the training process. A deep CNN architecture \cite{cc:86} typically consists of five convolutional layers and three fully connected layers, followed by a softmax classifier, and contains more than 60 million parameters. Even deeper networks, which achieve better performance, contain even more parameters. Training such large models on small datasets can lead to severe overfitting, even when overfitting prevention techniques are applied.

A common solution for training deep CNNs on small datasets is transfer learning, where the classifier layer of a pre-trained CNN is removed and fine-tuned on the target dataset. Transfer learning, which enables knowledge transfer from related tasks, has gained increasing attention in recent years. Several pre-trained CNN architectures have been proposed, including VGG, ResNet, Inception, and EfficientNet. In this study, we employ ResNet50 and InceptionV3, fine-tuning them on the palm fruit dataset for fruit maturity classification.

\subsection{ResNet50}
Instead of constructing a CNN from scratch, this study utilizes a pre-trained ResNet50 model. ResNet50 is a residual learning-based convolutional neural network with 50 layers. Residual learning has proven effective in addressing the vanishing gradient issue in very deep networks (those exceeding twenty layers) \cite{hcrt:94}. Compared to other architectures, ResNet models offer an additional advantage: their performance does not degrade as the network depth increases. The architecture of ResNet is illustrated in Figure \ref{ResNet}. The key innovation introduced by ResNet is the use of skip connections (shortcuts) to facilitate information flow across layers. ResNet begins with a single convolutional layer followed by max pooling. It then consists of four sequential layers, each with varying filter sizes, all utilizing $3 \times 3$ convolution operations. Additionally, after every two convolutional layers, a skip connection bypasses the intermediate layer, allowing the model to learn residual mappings instead of direct feature transformations. This concept defines the essence of ResNet models. These skipped connections, known as identity shortcut connections, are implemented using residual blocks. The residual block in ResNet is expressed as $H(x) = F(x)+x$, where $x$ is input layer, $H(x)$ is output layer, and $F(x)$ represents the residual mapping function. Residual blocks are effective when the input and output dimensions are identical, ensuring smooth information propagation. Each ResNet50 block consists of three layers. The initial layers of the ResNet architecture resemble those of GoogleNet, incorporating a $7 \times 7$ convolution operation followed by max pooling.

\subsection{InceptionV3}
The GoogLeNet network introduced a novel approach by integrating multiple Inception modules to form a Network-in-Network topology, as illustrated in Figure \ref{incpt}. The resulting network consists of 22 layers and achieved a winning top-5 error rate of 6.66\%. The significance of this architecture becomes evident when considering that it achieved such high accuracy with an order of magnitude fewer parameters compared to other models. The Inception module performs convolutions with multiple filter sizes on the input, applies max pooling, and concatenates the results before passing them to the next Inception module. A key innovation in this architecture is the introduction of $1 \times 1$ convolution operations, which significantly reduce the number of parameters, improving computational efficiency without sacrificing performance.

\begin{figure*}[t]
\centering
\includegraphics[width=0.98\textwidth]{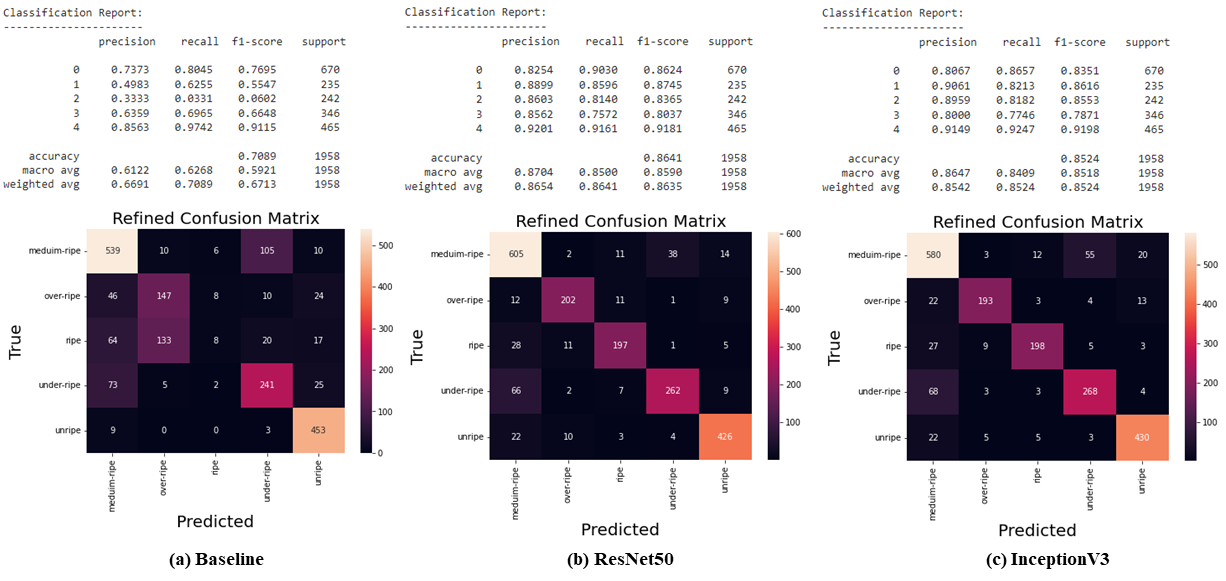} 
\caption{Classification report and confusion matrix, where classes 0, 1, 2, 3, and 4 correspond to medium-ripe, over-ripe, ripe, under-ripe, and unripe stages, respectively. Figures (a), (b), and (c) represent the results for the baseline model, ResNet50, and InceptionV3, respectively.}
\label{fig_baserc}
\end{figure*}

\begin{figure*}[t]
\centering
\includegraphics[width=0.98\textwidth]{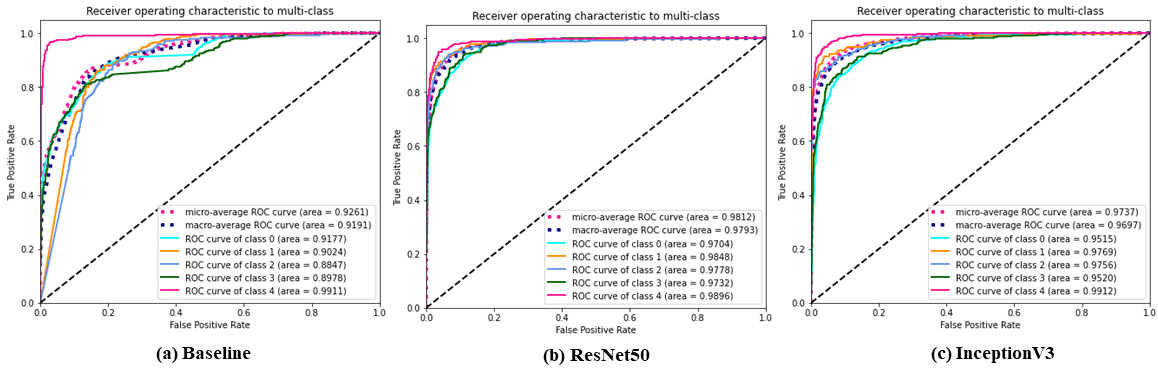} 
\caption{ROC curves using the micro-average method, macro-average method, and individual classes. Figures (a), (b), and (c) represent the results for the baseline model, ResNet50, and InceptionV3, respectively.}
\label{fig_rocs}
\end{figure*}

\section{Experiments}
We trained three models, a shallow CNN, ResNet50, and InceptionV3, for the palm fruit ripeness classification task. We utilized pre-trained parameters of ResNet50 and InceptionV3 for transfer learning. All experiments were conducted on Google Colab using Python 3, as Google Colab provides free GPUs for training. The dataset was downloaded from IEEE DataPort, and all images were reorganized according to their ripeness levels. The dataset was then processed and prepared for CNN architectures using the following steps:
\begin{itemize}
    \item Zip the dataset and upload it to Google Drive.
    \item Unzip the dataset on Google Colab for use in training. 
    \item Split the dataset into training (80\%) and testing (20\%) folders randomly.
    \item Use the \textit{flow\_from\_directory} method to load images directly from folders. Images were preprocessed using \textit{ImageDataGenerator} and then iterated through the generators with the \textit{next} function to obtain the data in the required X, y format for modeling. During preprocessing, images were rescaled to a range of 0-1 using a factor of 1/255. The \textit{class\_mode} parameter was set to categorical, meaning that labels were one-hot encoded.
\end{itemize}
After splitting the dataset into training and testing folders, 7,823 images were used for training, while 1,958 images were reserved for testing. Using the testing dataset, the shallow CNN, ResNet50, and InceptionV3 achieved final classification accuracies of 77.5\%, 85.9\%, and 84.5\%, respectively, for maturity classification.

\subsection{Evaluation Metrics}
The performance of the classification models was evaluated using accuracy (ACC), F1 score, confusion matrix, and the Area Under the Curve (AUC) of the Receiver Operating Characteristic (ROC) curve. Accuracy is defined in Equation (1):
\begin{equation}
    ACC=TP/N
\end{equation}
where $TP$ represents the number of correctly classified images, and $N$ is the total number of images in the test dataset. 

The F1 score combines precision and recall into a single metric using the harmonic mean, where the best score is 1 and the worst is 0. F1 score is defined in equation (2). Precision measures the accuracy of positive predictions and is defined as $TP/(TP+FP)$, where $TP$ refers to a case that belongs to class $x$ and is correctly predicted as $x$, while $FP$ refers to a case that does not belong to class $x$ but is incorrectly predicted as $x$. Recall measures the fraction of actual positive cases that are correctly identified and is defined as $TP/(TP+FN)$, where $FN$ refers to a case belonging to class $x$ that is incorrectly predicted as another class. A confusion matrix provides insight into how well models distinguish between different classes, helping to identify which classes are most easily confused. For each model, we used \textit{sklearn.metrics} to generate the confusion matrix and a classification report that includes precision, recall, and F1 score.
\begin{equation}
    F1\: score = \frac{2\times(Recall \times Precision)} {(Recall + Precision)}
\end{equation}
An ROC curve \cite{hcrt:92} is a graphical representation of a classification model’s performance across all classification thresholds. It visually illustrates model performance, while AUC (Area Under the Curve) quantifies the overall classification ability. AUC values range from 0 to 1, where AUC = 1 indicates a perfect model with 100\% correct predictions, and AUC = 0 represents a completely incorrect model with 100\% misclassifications. AUC is desirable because it is scale-invariant and classification-threshold-invariant. The ROC curve is generated using the true positive rate (TPR) and false positive rate (FPR). Both micro-average and macro-average methods were used to compute AUC values in our evaluation.

\begin{table*}
\centering
\begin{tabular}{|cc|ccc|}
\hline
\multicolumn{2}{|c|}{}                                          & \multicolumn{3}{c|}{Model}                                                                                           \\ \cline{3-5} 
\multicolumn{2}{|c|}{\multirow{-2}{*}{Evaluation Metrics}}      & \multicolumn{1}{c|}{Baseline} & \multicolumn{1}{c|}{ResNet50}                       & InceptionV3                    \\ \hline
\multicolumn{2}{|c|}{Accuracy}                                  & \multicolumn{1}{c|}{0.7089}   & \multicolumn{1}{c|}{\cellcolor[HTML]{C0C0C0}0.8641} & 0.8524                         \\ \hline
\multicolumn{1}{|c|}{}                           & Mic-average  & \multicolumn{1}{c|}{0.9261}   & \multicolumn{1}{c|}{\cellcolor[HTML]{C0C0C0}0.9812} & 0.9737                         \\ \cline{2-5} 
\multicolumn{1}{|c|}{}                           & Mac\_average & \multicolumn{1}{c|}{0.9191}   & \multicolumn{1}{c|}{\cellcolor[HTML]{C0C0C0}0.9793} & 0.9697                         \\ \cline{2-5} 
\multicolumn{1}{|c|}{}                           & Medium ripe  & \multicolumn{1}{c|}{0.9177}   & \multicolumn{1}{c|}{\cellcolor[HTML]{C0C0C0}0.9704} & 0.9515                         \\ \cline{2-5} 
\multicolumn{1}{|c|}{}                           & Over ripe    & \multicolumn{1}{c|}{0.9024}   & \multicolumn{1}{c|}{\cellcolor[HTML]{C0C0C0}0.9848} & 0.9769                         \\ \cline{2-5} 
\multicolumn{1}{|c|}{}                           & Ripe         & \multicolumn{1}{c|}{0.8847}   & \multicolumn{1}{c|}{\cellcolor[HTML]{C0C0C0}0.9778} & 0.9756                         \\ \cline{2-5} 
\multicolumn{1}{|c|}{}                           & Under ripe   & \multicolumn{1}{c|}{0.8978}   & \multicolumn{1}{c|}{\cellcolor[HTML]{C0C0C0}0.9732} & 0.9520                         \\ \cline{2-5} 
\multicolumn{1}{|c|}{\multirow{-7}{*}{AUC}}      & Unripe       & \multicolumn{1}{c|}{0.9911}   & \multicolumn{1}{c|}{0.9896}                         & \cellcolor[HTML]{C0C0C0}0.9912 \\ \hline
\multicolumn{1}{|c|}{}                           & Mic-average  & \multicolumn{1}{c|}{0.6713}   & \multicolumn{1}{c|}{\cellcolor[HTML]{C0C0C0}0.8635} & 0.8524                         \\ \cline{2-5} 
\multicolumn{1}{|c|}{}                           & Mac\_average & \multicolumn{1}{c|}{0.5921}   & \multicolumn{1}{c|}{\cellcolor[HTML]{C0C0C0}0.8590} & 0.8518                         \\ \cline{2-5} 
\multicolumn{1}{|c|}{}                           & Medium ripe  & \multicolumn{1}{c|}{0.7695}   & \multicolumn{1}{c|}{\cellcolor[HTML]{C0C0C0}0.8624} & 0.8351                         \\ \cline{2-5} 
\multicolumn{1}{|c|}{}                           & Over ripe    & \multicolumn{1}{c|}{0.5547}   & \multicolumn{1}{c|}{\cellcolor[HTML]{C0C0C0}0.8745} & 0.8616                         \\ \cline{2-5} 
\multicolumn{1}{|c|}{}                           & Ripe         & \multicolumn{1}{c|}{0.0602}   & \multicolumn{1}{c|}{0.8365}                         & \cellcolor[HTML]{C0C0C0}0.8553 \\ \cline{2-5} 
\multicolumn{1}{|c|}{}                           & Under ripe   & \multicolumn{1}{c|}{0.6648}   & \multicolumn{1}{c|}{\cellcolor[HTML]{C0C0C0}0.8037} & 0.7871                         \\ \cline{2-5} 
\multicolumn{1}{|c|}{\multirow{-7}{*}{F1 score}} & Unripe       & \multicolumn{1}{c|}{0.9115}   & \multicolumn{1}{c|}{0.9181}                         & \cellcolor[HTML]{C0C0C0}0.9198 \\ \hline
\end{tabular}
\caption{Performance comparisons of the baseline, ResNet50, and InceptionV3 models using accuracy, AUC, and F1 score. The highest values among the three models for different methods (classes) are highlighted in grey.}
\label{table1}
\end{table*}

\subsection{Baseline Model}
A classifier based on a simple CNN architecture was trained as our baseline model. This model consists of two convolutional layers, each followed by MaxPooling2D and Dropout layers. A flatten layer is then used to transform the extracted features into a one-dimensional vector, which is followed by a dense layer and another Dropout layer. The final dense layer with a softmax activation function computes a multinomial probability distribution. The architecture summary of the baseline model is shown in Figure \ref{fig_base}. The model was trained on 7,823 images with a learning rate of 0.0001, using categorical cross-entropy as the loss function and a batch size of 64. A total of 1,958 images was used to evaluate the model’s performance. After 40 epochs, the model achieved an accuracy of 79.04\% on the training dataset and 70.89\% on the test dataset.

The classification report and confusion matrix for this model are shown in Figure \ref{fig_baserc}(a). The model demonstrated high accuracy in distinguishing between ripe and unripe palm fruits, with only 12 unripe fruit bunches being misclassified (3 classified as under-ripe and 9 classified as medium-ripe). The F1 score for the unripe class (0.9115) is significantly higher than for other classes (all below 0.8). The ROC curve, shown in Figure \ref{fig_rocs}(a), visually supports this conclusion. However, the model exhibited low accuracy in correctly classifying palm fruits as ripe. Most ripe palm fruit bunches were misclassified as over-ripe or medium-ripe. Despite this, the misclassification is not a major issue during the harvesting process, as the oil quality and quantity produced in these stages do not differ significantly.
\begin{figure*}[t]
\centering
\includegraphics[width=0.9\textwidth]{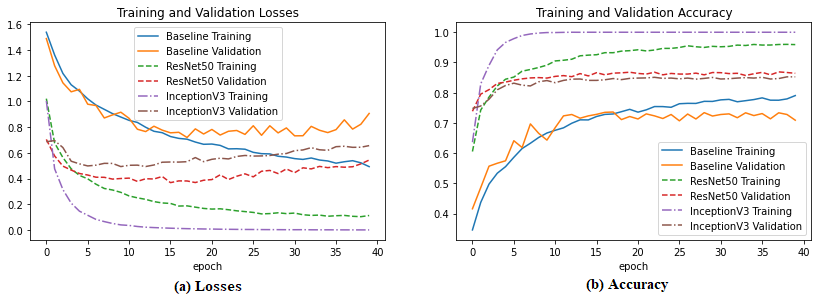} 
\caption{ Training and validation loss and accuracy trends for the baseline model, ResNet50, and InceptionV3 during the training phase. Figure (a) illustrates the training and validation loss, while Figure (b) depicts the training and validation accuracy. }
\label{fig_loss}
\end{figure*}

\subsection{ResNet50 Model}
The ResNet50 model was selected as one of the deep CNN architectures for our classification task. We imported the pre-trained ResNet50 model from \textit{Keras} applications and applied the following fine-tuning settings:
\begin{itemize}
\item \textit{weights='imagenet'} – The model utilized the weights learned from training on the ImageNet dataset.
\item \textit{include\_top=False} – This allowed us to customize the input and output layers according to our dataset.
\item \textit{layer.trainable=False} – This ensured that the pre-trained layers retained their learned weights, reducing computational cost and preventing redundant training.
\item After flattening the pre-trained layers from ResNet50, we added two dense layers. The first dense layer contained 16 units with a ReLU activation function, followed by a Dropout layer to mitigate overfitting. The final dense output layer utilized a softmax activation function with five output neurons, corresponding to the five ripeness classes in our dataset.
\end{itemize}

This model was trained on the same dataset using the Adam optimizer, a learning rate of 0.001, and categorical cross-entropy as the loss function. After 40 epochs, it achieved an accuracy of 95.91\% on the training set and 86.41\% on the test set. The model performed well in classifying palm fruit bunches into the correct maturity stages. The F1 scores for all maturity classes were above 0.8, as shown in Figure \ref{fig_baserc}(b). However, the classifier had the lowest F1 score for the under-ripe class. Upon examining the confusion matrix, we observed that a significant number of under-ripe palm bunches were misclassified as medium-ripe. This misclassification may have been influenced by the presence of green protective bags covering some bunches. The ROC curve for this model is shown in Figure \ref{fig_rocs}(b). The AUC values under the micro-average method, macro-average method, and individual classes were all close to 1, indicating that this model is a highly effective classifier for palm bunch maturity classification.

\subsection{InceptionV3 Model}
The InceptionV3 architecture was another model fine-tuned for the palm bunch maturity classification task. The pre-trained model was imported from \textit{Keras} applications, utilizing pretrained parameters from the ImageNet dataset. We followed a process similar to the ResNet50 model, modifying only the input and output layers to fit our dataset. After flattening the pre-trained InceptionV3 layers, we added a dense output layer with five units and a softmax activation function to compute the multinomial probability distribution for the five maturity classes.

The model was trained using the same dataset split settings. It was optimized using Adam, with a learning rate of 0.0001, and categorical cross-entropy as the loss function. After 40 epochs, it achieved an accuracy of 100\% on the training set and 85.24\% on the test set. The performance of this model was fairly strong for the classification task. As shown in Figure \ref{fig_baserc}(c) the F1 scores for all classes were above 0.8, except for the medium-ripe class. The confusion matrix exhibited a pattern similar to the ResNet50 model, where a large number of under-ripe palm bunches were misclassified as medium-ripe. Similar to the ROC curve of the ResNet50 model, shown in Figure \ref{fig_rocs}(c), the AUC values for this model were also very close to 1, further indicating its effectiveness in palm bunch maturity classification.

\subsection{Comparisons of Models}
The summary of accuracy, AUC, and F1 score using different methods for the baseline model, ResNet50, and InceptionV3 is presented in Table \ref{table1}. The highest values among the three models for different methods or classes are highlighted in grey. ResNet50 outperformed the other models in most evaluation metrics. Although InceptionV3 achieved the highest AUC for the unripe class and the highest F1 scores for the ripe and unripe classes, its overall performance was not significantly better than that of ResNet50.

Figure \ref{fig_loss} illustrates the training and validation loss and accuracy trends for the baseline model, ResNet50, and InceptionV3 during training. The baseline model exhibited a slow decrease in loss, with validation accuracy peaking at approximately 70\% after 15 epochs and stabilizing thereafter. For the InceptionV3 model, the loss decreased the fastest, but validation accuracy stabilized after just 10 epochs. The training accuracy reached 100\% within a few epochs, yet the test accuracy remained around 80\%, indicating a clear overfitting problem. The ResNet50 model also showed a rapid decrease in loss, with validation accuracy stabilizing after six epochs. Throughout training, the accuracy difference between the training and validation sets remained small, suggesting better generalization. These results demonstrate that the proposed deep CNN models performed well, with ResNet50 achieving the best overall performance. 

\section{Discussion and Conclusion}
A machine vision framework for palm fruit maturity classification was proposed based on deep learning. A shallow CNN was used as the baseline model, while transfer learning with fine-tuning was applied to the classification task. Two pre-trained CNN models, ResNet50 and InceptionV3, were investigated. The study utilized a published dataset designed with a high degree of variation to reflect the challenges present in natural environments and fruit orchards. The best accuracy was achieved by the fine-tuned ResNet50 model, which attained 86.41\% for maturity classification.

This study demonstrated that a pre-trained CNN could achieve robust fruit classification without the need for image preprocessing to remove background noise or enhance illumination. By leveraging pre-trained models, high classification accuracy can be obtained within a few training epochs, as the feature extraction process is already well-optimized by the original architecture. However, further research is necessary. EfficientNet, another pre-trained model, was originally included in our study. Despite extensive parameter tuning and multiple training attempts, we observed that while training accuracy was very high, test accuracy remained below 50\%. Future work should investigate the reasons behind this poor generalization and explore ways to enhance its performance. Additionally, the maturity stages of palm fruit are influenced by fruit type. To improve classification performance, a type classification task will be incorporated to mitigate errors caused by different palm fruit varieties.

\section{References}
\nobibliography{references}

\bibentry{hcrt:83}.\\[.2em]
\bibentry{hcrt:84}.\\[.2em]
\bibentry{c:83}. \\[.2em]
\bibentry{hcrt:85}. \\[.2em]
\bibentry{hcrt:86}. \\[.2em]
\bibentry{hcrt:87}. \\[.2em]
\bibentry{hcrt:88}. \\[.2em]
\bibentry{cc:83}. \\[.2em]
\bibentry{cc:84}. \\[.2em]
\bibentry{hcrt:89}. \\[.2em]
\bibentry{hcrt:90}. \\[.2em]
\bibentry{cc:85}. \\[.2em]
\bibentry{hcrt:91}. \\[.2em]
\bibentry{hcrt:92}. \\[.2em]
\bibentry{hcrt:93}. \\[.2em]
\bibentry{cc:86}. \\[.2em]
\bibentry{hcrt:94}. \\[.2em]
\bibentry{cc:87}.

\end{document}